\documentclass[runningheads]{llncs}

\usepackage{eccv}
\usepackage{hyperref}

\newcommand{\cmark}{\ding{51}}
\newcommand{\xmark}{\ding{55}}

\usepackage[accsupp]{axessibility}  
\usepackage{wrapfig}
\usepackage{multirow}
\usepackage{multirow}

\usepackage{graphicx}
\usepackage{booktabs}
\usepackage[table]{xcolor}
\usepackage{threeparttable}
\usepackage{multirow}

\usepackage{graphicx}
\usepackage{booktabs}  
\usepackage{pifont}  

\usepackage{eccvabbrv}

\usepackage{graphicx}
\usepackage{booktabs}

\usepackage[accsupp]{axessibility}  

\usepackage{hyperref}

\usepackage{orcidlink}

\usepackage{marvosym}
\begin{document}

\title{ArcAD: Anomaly-Rectified Calibration for Cold-Start Supervised Anomaly Detection} 

\titlerunning{ArcAD}

\authorrunning{Han et al.}

\author{Ningning Han\inst{1,6}\thanks{Equal contribution.} \and
Lei Fan\inst{2,6}$^{\star} $ \and
Jia Guo\inst{3} \and
Yunkang Cao\inst{4} \and
Xiu Su\inst{5} \and
Feng Cao\inst{6} \and
Donglin Di\inst{1,6} \and
Tonghua Su\inst{1,7}}

\institute{Faculty of Computing, Harbin Institute of Technology, Harbin, China \\
\email{25B903152@stu.hit.edu.cn, thsu@hit.edu.cn} \and
University of New South Wales, Sydney, Australia \\
\email{lei.fan1@unsw.edu.au} \and
School of Biomedical Engineering, Tsinghua University, Beijing, China \\ 
\and School of Artificial Intelligence and Robotics, Hunan University, Changsha, China \\ %
  \and
Big data institute, Central South University, Changsha, China \\ 
\and DZ-Matrix, Beijing, China
 \\
 \and Chongqing Research Institute of HIT, Chongqing, China
}

\maketitle

\begingroup
\renewcommand{\thefootnote}{}
\footnotetext{Corresponding authors: Tonghua Su and Lei Fan.}         
\endgroup

\begin{abstract}
The deployment of Industrial Anomaly Detection (IAD) in real-world manufacturing frequently encounters a challenging cold-start bottleneck, in which limited normal samples fail to represent the full normal distribution and only a few anomalies are available. Under such a regime, existing methods struggle to form compact normal boundaries and fail to effectively exploit supervised signals from rare defects. To address this challenge, we propose Anomaly-Rectified Cold-start AD (ArcAD), a plug-and-play calibration framework for reconstruction-based IAD baselines. ArcAD follows a push–pull learning paradigm to construct a compact and discriminative normal boundary under data scarcity. On the one hand, ArcAD projects limited normal samples onto a hypersphere and pulls them into multiple compact clusters to maximize coverage of the normal manifold. On the other hand, it synthesizes pseudo-anomalies on the hypersphere and leverages real anomalies to push the boundary inward and sharpen anomaly discrimination.
Extensive experiments on MVTec-AD, VisA, Real-IAD, and MANTA demonstrate that ArcAD significantly outperforms state-of-the-art supervised and unsupervised methods in both single-class and multi-class settings under cold-start conditions. Code is available at: \url{https://github.com/LGC-AD/ArcAD}.

  \keywords{Anomaly Detection \and Hypersphere \and Cold Start }
\end{abstract}

\section{Introduction}
\label{sec:intro}

Industrial Anomaly Detection (IAD) seeks to achieve zero-defect manufacturing by identifying rare defects, such as scratches, cracks, or structural deformations~\cite{fan2025salvaging, jeong2023winclip, fan2025grainbrain, ling2025adnet}. The deployment of IAD systems in real-world applications frequently encounters a severe \textit{cold-start} bottleneck. Unlike standard unsupervised settings~\cite{ruff2021unifying, guo2025one} that assume abundant normal data, cold-start scenarios during the initial ramp-up phase of a new production line present a distinct data distribution. During this stage, the available normal samples fail to cover the full range of normal patterns~\cite{roth2022towards, rudolph2021same}. Meanwhile, only a small number of anomalies are collected during early system deployment.

\begin{figure*}[t]
    \centering
    \begin{minipage}[c]{0.61\linewidth}
        \centering
        \includegraphics[width=\linewidth]{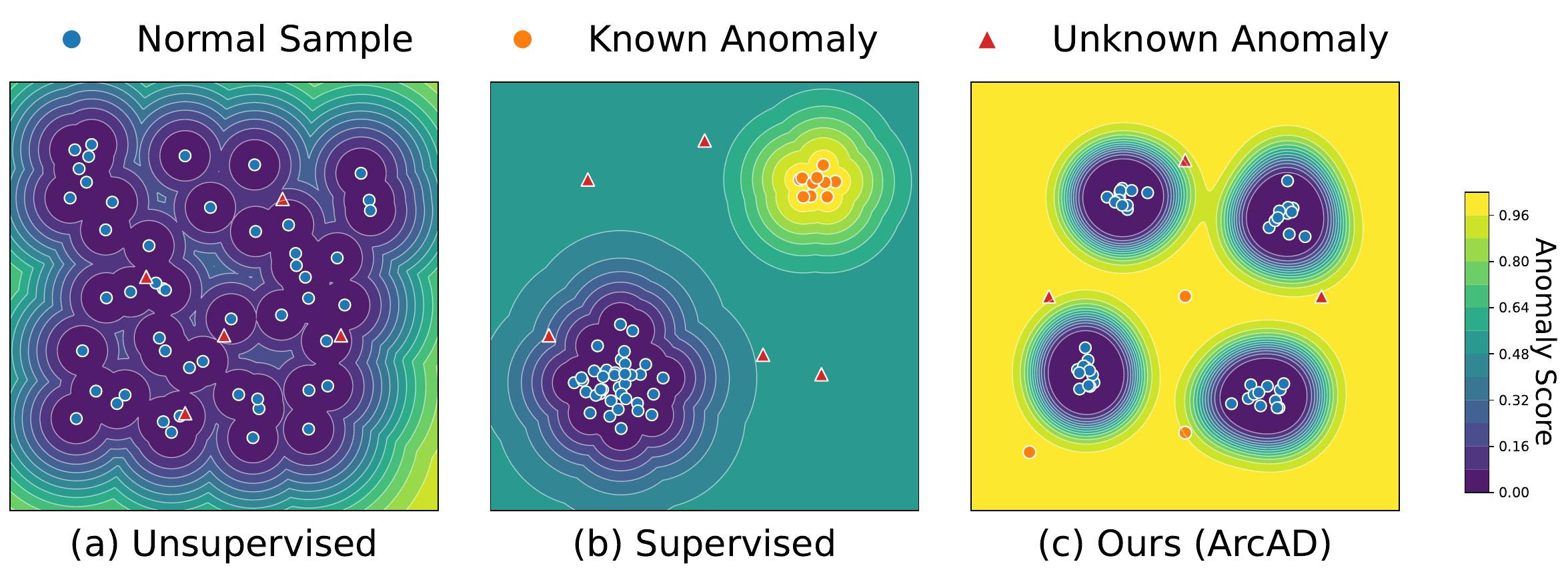}
    \end{minipage}
    \hfill
    \begin{minipage}[c]{0.38\linewidth}
        \centering
        \renewcommand{\arraystretch}{1.3} 
        
        \resizebox{\linewidth}{!}{
            \begin{tabular}{lccc}
                \toprule
                 & \textbf{Unsup.} & \textbf{Sup.} & \textbf{Ours} \\
                \midrule
                Data-efficient (Normal)    & \xmark & \xmark & \cmark \\
                Anomaly Utilization   & \xmark & \cmark &\cmark \\
                Generalizable (Unseen)  & \cmark & \xmark & \cmark \\
                \bottomrule
            \end{tabular}
        }
    \end{minipage}
    
    \caption{\textbf{Left:} Decision boundaries in cold-start scenarios. Unsupervised methods (a) yield loose manifolds and supervised methods (b) overfit to known anomalies, whereas ArcAD (c) forms compact and unified boundaries. \textbf{Right:} Paradigm comparison on data efficiency, anomaly utilization, and generalization.}
    \label{fig:motivation}
\end{figure*}

Recent supervised methods~\cite{ding2022catching, yao2023explicit, zhu2024anomaly} show promising performance when trained with annotated normal and abnormal samples. However, in cold-start scenarios, these methods are prone to overfitting when trained on only a few anomalies, leading to poor generalization to unseen defect patterns (see Fig.~\ref{fig:motivation}). 
Conversely, unsupervised approaches, particularly reconstruction-based methods~\cite{he2024mambaad, guo2025dinomaly,deng2022anomaly}, have established dominance in both single-class and multi-class settings~\cite{you2022unified}, primarily due to their strong ability to generalize to diverse and unseen anomaly patterns. 
These methods typically learn the distribution of normal data and identify instances that deviate from the learned normal manifold as anomalies.
While abundant data allows models to implicitly learn a continuous and robust normal manifold, data scarcity in cold-start scenarios leads to a fragmented and loosely bounded latent space. Moreover, by strictly relying on normal data, these unsupervised paradigms inherently underutilize the valuable guidance provided by the few available anomaly samples. Therefore, a pivotal question emerges: \textit{How can we leverage limited normal samples and rare anomalies to construct a compact normal boundary?}

To address this challenge, we propose Anomaly-Rectified Cold-start AD (ArcAD), a plug-and-play framework designed to enhance reconstruction-based models. ArcAD constructs a compact and discriminative normal boundary from two complementary perspectives: explicitly organizing normal samples into compact clusters and calibrating the boundary using anomaly signals. 
First, we project patch features onto a hypersphere, separating directional information from magnitude variations to establish a bounded geometric representation~\cite{defard2021padim}. These hyperspherical features are then modeled using the von Mises-Fisher (vMF) distribution~\cite{mardia2009directional}. To ensure that limited normal data adequately covers the latent manifold, we introduce Sinkhorn-based Prototype Modeling (SPM). By formulating feature clustering as an optimal transport problem, SPM mitigates biased feature aggregation and partitions normal embeddings into compact and uniform distributed clusters. 

Beyond modeling the normal manifold, ArcAD further leverages anomaly signals to explicitly rectify the normal boundary. To alleviate the scarcity of anomalies, we design a prototype-restricted anomaly synthesis strategy, which generates synthetic anomalies directly on the hypersphere by filtering candidate samples against normal prototypes. These synthesized samples, together with the few genuine defects, drive Defect-Guided Calibration (DGC). This module employs a contrastive objective that pulls anomalies together while pushing them away from their nearest normal prototypes. This process explicitly reinforces the compactness and discriminative power of the learned normal manifold.
We summarize our contributions as follows:

\begin{itemize}
\item We propose ArcAD, a generic plug-and-play calibration framework for reconstruction based models under cold-start scenarios. 

\item We introduce Sinkhorn-based Prototype Modeling (SPM) to organize limited normal samples into compact and uniform clusters on the hypersphere. 
\item We propose Defect-Guided Calibration (DGC), which introduces a prototype-restricted synthesis strategy to generate pseudo-anomalies, thereby leveraging both real and synthesized defects to explicitly rectify the normal boundary in the latent space.
\item Extensive experiments on four datasets~\cite{bergmann2019mvtec, zou2022spot, wang2024real, fan2025manta} demonstrate that ArcAD consistently enhances state-of-the-art baselines such as ReContrast~\cite{guo2023recontrast}, RD4AD~\cite{deng2022anomaly}, and Dinomaly~\cite{guo2025dinomaly}. Specifically, under the challenging multi-class setting of the Real-IAD dataset~\cite{wang2024real}, ArcAD achieves Image-level AUROC gains of +2.2\%, +8.9\%, and +3.7\% for these models, respectively.
\end{itemize} 


\section{Related Work}
\paragraph{Industrial Anomaly Detection.}
Depending on the availability of training data, existing IAD approaches can be broadly categorized into unsupervised and supervised AD methods. 
Unsupervised AD typically follows a one-class classification paradigm, including reconstruction-based~\cite{guo2025dinomaly, zhang2023exploring, he2024mambaad, guo2023recontrast} and embedding-based methods~\cite{yao2024hierarchical, mcintosh2023inter, roth2022towards}. These approaches learn the distribution of normal samples and detect anomalies as deviations during inference. However, under cold-start scenarios, limited normal samples often lead to loosely defined latent boundaries, significantly degrading detection performance. 
Supervised AD~\cite{yao2023explicit, pang2019deep, liscale} instead leverages the prior knowledge derived from observed anomalies to mitigate false positives. For instance, DRA~\cite{ding2022catching} learns representations of observed anomalies, pseudo-anomalies, and residual anomalies to enhance anomaly discrimination. Similarly, DPDL~\cite{wang2025distribution} confines normal samples within a compact feature space and employs the Schr\"{o}dinger bridge to guide anomalous samples towards normal distribution. However, these methods tend to overfit to the limited observed anomalies, which restricts their generalization to unseen defect patterns.

\paragraph{Hypersphere Learning.}
Constraining representations onto a unit hypersphere has emerged as an effective strategy for enhancing feature uniformity and embedding alignment~\cite{wang2020understanding,davidson2018hyperspherical,fu2025beyond}. To statistically characterize these hyperspherical representations, the von Mises-Fisher (vMF) distribution~\cite{mardia2009directional} provides a principled probabilistic foundation and has been widely adopted in clustering~\cite{qin2016online, gopal2014mises}, semantic segmentation~\cite{hwang2019segsort}, and face recognition~\cite{xu2023probabilistic,kobayashi2021t}. 
In addition, hyperspherical representations have also been explored in generative modeling to constrain latent spaces and improve generation fidelity~\cite{ke2025hyperspherical, loshchilovngpt}. More recently, hyperspherical modeling has demonstrated advantages in out-of-distribution detection~\cite{ming2022exploit, lu2024learning, li2024learning} by naturally providing a bounded latent space.

Motivated by these geometric properties, we introduce ArcAD, which models features on a unit hypersphere to organize limited normal samples into a compact manifold while leveraging rare anomalies to rectify the normal boundary.


\section{Methodology}

\subsection{Overview}
\paragraph{Cold-Start Setting.} 
We consider industrial anomaly detection under a realistic \textit{cold-start} setting, where only limited normal samples and a few genuine defect samples are available for training. Let $\mathcal{X} = \mathcal{X}_N \cup \mathcal{X}_A$ denote the training set, where $\mathcal{X}_N = \{x_i\}$ represents normal images and $\mathcal{X}_A = \{x_j\}$ represents a small set of genuine defect samples,  with $|\mathcal{X}_N| \gg |\mathcal{X}_A|$ (\textit{e.g.,} $9{:}1$). 
The overall training data is limited to reflect early-stage deployment, while normal samples strictly outnumber anomalies.
For evaluation, we define the test set as $\mathcal{X}_{test} = \mathcal{X}_{test, N} \cup \mathcal{X}_{test, A}$. The objective is to learn a compact and discriminative boundary for normal data under the cold-start regime.

\paragraph{Framework Overview.}
ArcAD is built upon reconstruction-based IAD frameworks~\cite{guo2025dinomaly,guo2023recontrast, deng2022anomaly} with an encoder $\mathcal{E}(\cdot)$, a bottleneck $\mathcal{B}(\cdot)$, and a decoder $\mathcal{D}(\cdot)$. Given an input image $x_i\in \mathcal{X}_N$, the encoder and bottleneck map it to a latent representation $z_i = \mathcal{B}(\mathcal{E}(x_i))$. 
The decoder then reconstructs representations from the latent space. The optimization objective can be formulated as:
\begin{equation}
    \mathcal{L}_{recon} = d( \mathcal{E}(x_i), \, \mathcal{D}(\mathcal{B}(\mathcal{E}(x_i)))),
    \label{eq:recon}
\end{equation}
where $d(\cdot,\cdot)$ denotes a distance function (\textit{e.g.,} mean squared error or cosine distance) measuring the discrepancy between the original and reconstructed features.
While reconstruction-based loss implicitly captures the normal manifold, under cold-start conditions, the limited normal data can lead to a biased and loosely bounded latent structure, yielding an imprecise normal boundary.

\begin{figure*}[t] 
    \centering
    \includegraphics[width=0.98\textwidth]{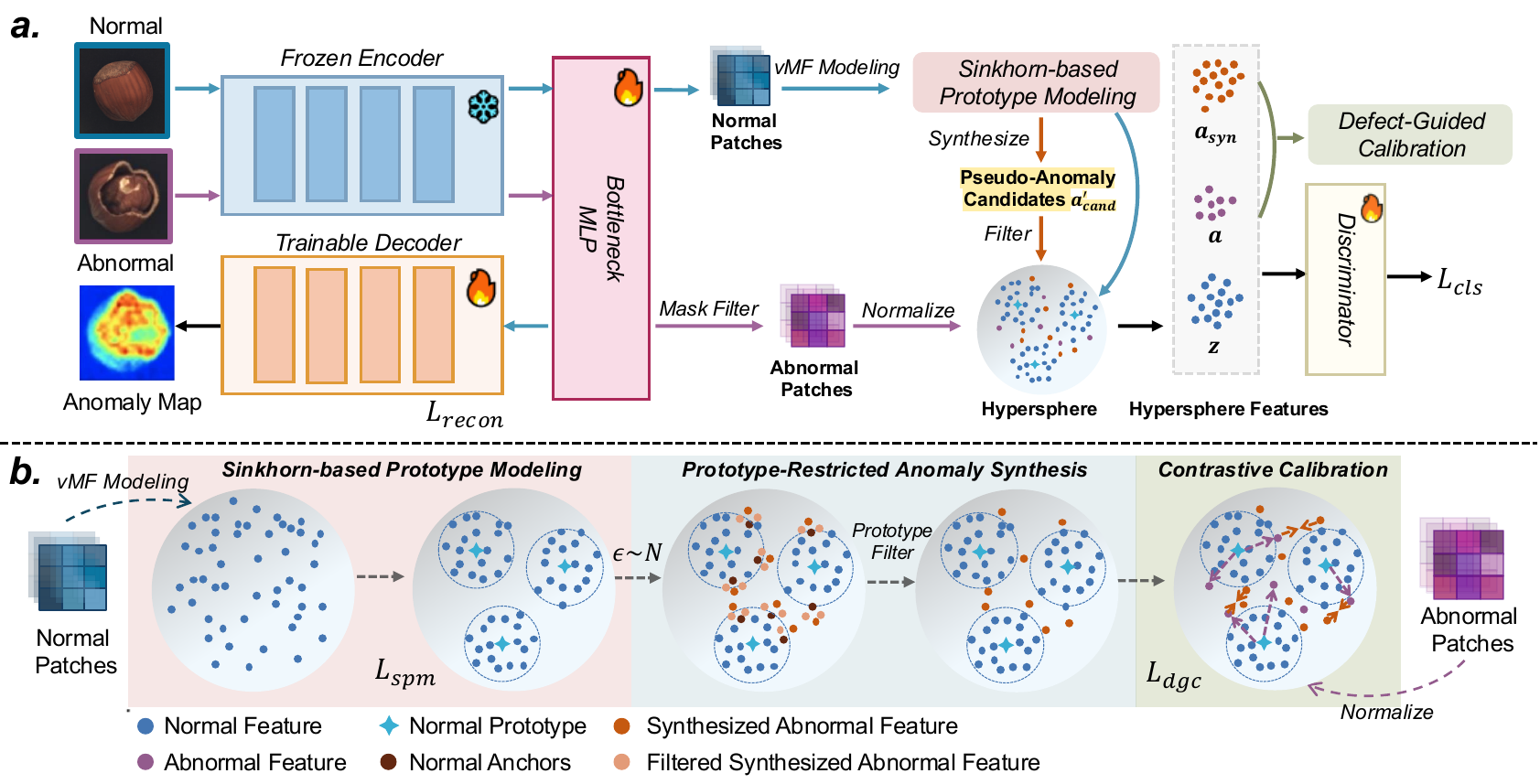}
    \caption{\textbf{Overview of the proposed ArcAD.} (a) Integration of the ArcAD framework into a standard reconstruction-based architecture. (b) ArcAD explicitly calibrates the latent feature distribution through Sinkhorn-based Prototype Modeling (SPM) and Defect-Guided Calibration (DGC). 
    }
    \label{fig:main_framework} 
\end{figure*}

To mitigate this issue, we propose Anomaly-Rectified Cold-start AD (ArcAD), a plug-and-play calibration framework that operates on the latent space without changing the reconstruction backbone (see Fig.~\ref{fig:main_framework}). 
ArcAD consists of two core components: Sinkhorn-based Prototype Modeling (SPM), which organizes limited normal samples to form a compact and well-covered normal structure, and Defect-Guided Calibration (DGC), which refines the normal boundary by leveraging both synthesized pseudo-anomalies and real anomalies.


\subsection{Sinkhorn-based Prototype Modeling (SPM)} 

\paragraph{vMF Modeling.} 
Compared to the unbounded Euclidean space, the hypersphere provides a compact manifold~\cite{wang2020understanding, loshchilovngpt}, which naturally constrains the feature distribution and facilitates the formation of compact normal structures. Prior studies~\cite{rippel2021gaussian, defard2021padim} show that deep representations of normal patterns naturally form compact clusters that closely follow a Gaussian-like distribution in the latent space. Accordingly, we explicitly model the hyperspherical features using the von Mises-Fisher (vMF) distribution~\cite{mardia2009directional, ming2022exploit}, the hyperspherical analogue of the Gaussian distribution. 

Let $\{z_i \in \mathbb{R}^D\}_{i=1}^{N_p}$ denote the latent patch embeddings extracted by the bottleneck for an input image, where $N_p = H_f \times W_f$ is the number of spatial patches. We project each feature $z_i$ onto the unit hypersphere via $\ell_2$ normalization, \textit{i.e.,} $\mathbf{z}_i = z_i / \|z_i\|_2$, such that $\mathbf{z}_i \in \mathbb{S}^{D-1}$. The probability density function for a normalized patch feature $\mathbf{z}$ following a vMF distribution is defined as:
\begin{equation}
    f(\mathbf{z}; \boldsymbol{\mu}, \kappa) = C_D(\kappa) \exp(\kappa \boldsymbol{\mu}^\top \mathbf{z}),
    \label{eq:vmf_pdf}
\end{equation}
where $\boldsymbol{\mu} \in \mathbb{S}^{D-1}$ denotes the mean direction representing the cluster prototype, $\kappa \ge 0$ is the concentration parameter controlling the compactness around $\boldsymbol{\mu}$, and $C_D(\kappa)$ is the normalization constant.


\paragraph{Sinkhorn-based Clustering.} 
While the vMF distribution characterizes directional concentration on the hypersphere, normal data in practice often exhibit multiple modes due to diverse textures and structures, especially in multi-class settings. Therefore, instead of modeling the normal distribution with a single prototype, we learn multiple prototypes to capture diverse normal patterns.
Under cold-start conditions, limited normal samples may lead to a biased feature space~\cite{roth2022towards, park2020learning}, where prototypes collapse into a few dense regions. To encourage uniform prototype learning, we formulate the feature-to-prototype assignment as a balanced optimal transport problem.

Specifically, we initialize $K$ prototypes $\mathcal{U}=\{\boldsymbol{\mu}_k\}_{k=1}^K$ on the hypersphere using $K$-means with cosine distance. 
Given a batch with $N_{\text{batch}}$ normalized patch features, we seek a soft assignment matrix $Q \in \mathbb{R}^{N_{\text{batch}} \times K}$, where $Q_{ik}$ denotes the assignment weight between feature $\mathbf{z}_i$ and prototype $\boldsymbol{\mu}_k$.
The assignment is optimized to favor high feature--prototype similarity while satisfying an equipartition constraint. Concretely, each feature is fully assigned across prototypes (row sums equal $1$), and each prototype receives equal total mass (column sums equal $N_{\text{batch}}/K$).

To compute $Q$, we solve this optimal transport problem using the iterative Sinkhorn-Knopp algorithm~\cite{cuturi2013sinkhorn}. We compute the pairwise cosine similarity matrix $S \in \mathbb{R}^{N_{\text{batch}} \times K}$ with $S_{ik} = \mathbf{z}_i^\top \boldsymbol{\mu}_k$, and initialize the exponential similarity matrix as $M = \exp(S / \epsilon)$, where $\epsilon$ is a temperature parameter. The algorithm then derives an entropically regularized solution formulated as:
\begin{equation}
    Q = \text{diag}(\mathbf{u}) M \text{diag}(\mathbf{v}),
\end{equation}
where $\mathbf{u} \in \mathbb{R}^{N_{\text{batch}}}$ and $\mathbf{v} \in \mathbb{R}^{K}$ are non-negative scaling vectors. 
The assignment and prototype update follow an Expectation-Maximization (EM)-style procedure~\cite{moon1996expectation}. In the E-step, we infer $Q$ with fixed network parameters and prototypes. In the M-step, we update prototypes $\boldsymbol{\mu}_k$ using the exponential moving average (EMA)~\cite{he2020momentum} to ensure stable evolution on the hypersphere.

With the optimal transport plan $Q$ acting as the soft assignment weight, we formulate $\mathcal{L}_{spm}$ as a cross-entropy objective to align the network predictions: 
\begin{equation}
    \mathcal{L}_{spm} = - \frac{1}{N_{\text{batch}}} \sum_{i=1}^{N_{\text{batch}}} \sum_{k=1}^{K} Q_{ik} \log \frac{\exp(\mathbf{z}_i^\top \boldsymbol{\mu}_k / \tau)}{\sum_{j=1}^{K} \exp(\mathbf{z}_i^\top \boldsymbol{\mu}_j / \tau)},
    \label{eq:spm_loss}
\end{equation}
where $\tau=1/\kappa$ is a temperature hyperparameter. By optimizing $\mathcal{L}_{spm}$, we enforce the normal representations to form compact clusters within the latent space. It is crucial to note that the Sinkhorn mechanism strictly enforces uniformity to ensure no single prototype dominates the mass assignments, effectively preventing biased learning toward high-density regions. Concurrently, the comprehensive coverage of the normal manifold is achieved by the large capacity of the multi-cluster prototypes ($K$). Together, these properties establish a highly robust and discriminative normal boundary.



\subsection{Defect-Guided Calibration (DGC)}
Building upon the normal prototypes learned by SPM, we next use anomaly signals to explicitly calibrate the normal boundary. In particular, real anomalies serve as negative samples that are pushed away from the normal prototypes, enforcing a margin between anomalous samples and the normal distribution.

In the cold-start regime, however, only a few real anomalies are available, which provide insufficient support to shape a stable anomalous region. To alleviate this scarcity, we introduce a prototype-restricted anomaly synthesis strategy to expand the negative set near the boundary. Unlike image-level augmentation (\textit{e.g.,} Cutpaste~\cite{li2021cutpaste} or DRAEM~\cite{zavrtanik2021draem}) or feature-level augmentation in unbounded Euclidean space (\textit{e.g.,} SimpleNet~\cite{liu2023simplenet}), we perform perturbations directly on the hypersphere so that synthesized samples remain within a bounded geometry and avoid drifting into trivial outliers.

\paragraph{Prototype-Restricted Anomaly Synthesis.} 
Given a normal feature $\mathbf{z} \in \mathbb{S}^{D-1}$ as an anchor, we generate a candidate set of perturbed features by injecting Gaussian noise $\boldsymbol{\xi} \sim \mathcal{N}(0, \sigma^2 \mathbf{I})$ and re-projecting the resulting features onto the unit hypersphere:
\begin{equation}
\mathbf{a}'_{cand} = (\mathbf{z} + \boldsymbol{\xi})/\|\mathbf{z} + \boldsymbol{\xi}\|_2.    
\end{equation}
However, blindly using all candidates may corrupt the normal manifold, as some candidates can remain close to normal regions. To enforce a discriminative margin, we filter the candidates using the learned normal prototypes $\mathcal{U} = \{\boldsymbol{\mu}_k\}_{k=1}^K$ as geometric constraints. Specifically, we select the candidate that minimizes its maximum similarity to the prototypes:
\begin{equation}
\mathbf{a}_{syn} =
\arg\min_{\mathbf{a}'_{cand}\in \mathcal{A}_{cand}'}
\max_{\boldsymbol{\mu}_k \in \mathcal{U}}
(\mathbf{a}'_{cand})^\top \boldsymbol{\mu}_k.
\label{eq:hard_sample_selection}
\end{equation}
In practice, for each normal anchor feature, we generate $N_{cand}=5$ independent noise perturbations to form the candidate set $\mathcal{A}'_{cand}$, from which exactly one pseudo-anomaly is selected.

\paragraph{Contrastive Calibration.} 
We calibrate the boundary with two complementary objectives. First, for real anomaly features, we penalize the maximum similarity to normal prototypes, pushing anomalies away from the normal distribution. Second, we align each real anomaly with synthesized pseudo-anomalies, encouraging a coherent anomalous region near the boundary.

Formally, given an abnormal image $\mathbf{x}_j \in \mathcal{X}_A$, we extract its patch-level features as a set $\mathcal{A} = \{\mathbf{a}_i\}_{i=1}^{N_p} \subset \mathbb{R}^D$. Guided by its down-sampled mask, we isolate and $\ell_2$-normalize anomalous patch embeddings onto the unit hypersphere, forming the anomaly set:
\begin{equation}
\hat{\mathcal{A}} = \{ \mathbf{a}_i / \|\mathbf{a}_i\|_2 \mid \mathbf{a}_i \in \mathcal{A}, m_i = 1 \}
\end{equation}
where $m_i \in \{0, 1\}$ is the $i$-th element of the ground-truth mask $M \in \{0, 1\}^{N_p}$.  

Let $\mathcal{A}_{syn}$ denote the set of all synthesized pseudo-anomalies generated from the normal anchors within a batch.
Integrating the two objectives, the defect-guided calibration loss $\mathcal{L}_{dgc}$ is formulated as:
\begin{equation}
    \mathcal{L}_{dgc} = \mathbb{E}_{\mathbf{a} \in \hat{\mathcal{A}}} \left( \max_{\boldsymbol{\mu}_k \in \mathcal{U}} (\mathbf{a}^\top \boldsymbol{\mu}_k) + \left( 1 - \frac{1}{|\mathcal{A}_{syn}|} \sum_{\mathbf{a}_{syn} \in \mathcal{A}_{syn}} \mathbf{a}^\top \mathbf{a}_{syn} \right) \right),
    \label{eq:calib_loss}
\end{equation}
where $|\mathcal{A}_{syn}|$ denotes the size of the synthesized anomaly set.
Consequently, these anomalies provide a repulsive signal that sharpens the normal boundary and leads to a more compact fit of the normal distribution.


\subsection{Optimization}
\paragraph{Training.}
We introduce a lightweight discriminator $f_{disc}(\cdot)$ (\textit{e.g.,} a 2-layer MLP) to act as a global structural regularizer rather than merely a probability predictor. Combined with the SPM and DGC modules, it further enforces a clear decision margin in the latent space while maintaining the compactness of the normal hyperspherical manifold. To achieve this while mitigating the class imbalance inherent in the cold-start setting, the discriminator is optimized using the Binary Focal Loss~\cite{lin2017focal}:
\begin{equation}
    \mathcal{L}_{cls} = \mathbb{E}_{\mathbf{e} \in \mathcal{Z} \cup \mathcal{A}_{syn} \cup \hat{\mathcal{A}}} \left[ \mathcal{L}_{FL}(f_{disc}(\mathbf{e}), y) \right],
\end{equation}
where $\mathcal{Z} = \{\mathbf{z}_i\}$ denotes the set of normal features, and $y \in \{0, 1\}$ is the corresponding binary label ($0$ for normal features, and $1$ for both real and synthetic anomalies).

During the training phase, the bottleneck network is optimized using the following comprehensive objective function:
\begin{equation}
    \mathcal{L}_{total} = \mathcal{L}_{recon} + \lambda_1 \mathcal{L}_{spm} + \lambda_2 \mathcal{L}_{dgc} + \lambda_3 \mathcal{L}_{cls},
    \label{eq:total_loss}
\end{equation}
where $\lambda_1, \lambda_2$, and $\lambda_3$ are hyperparameters balancing the contribution of each module. The decoder is updated using only the reconstruction loss ($\mathcal{L}_{recon}$) derived from normal samples, while the parameters of the discriminator are optimized solely via the classification loss ($\mathcal{L}_{cls}$). 

\paragraph{Inference.} ArcAD does not alter the standard reconstruction-based anomaly scoring mechanism. Instead, it regularizes the latent space during training. At inference time, anomaly scores are computed directly from the reconstruction-based score map~\cite{guo2025dinomaly,deng2022anomaly}.

\section{Experiments}
\subsection{Experimental Setup}

\paragraph{Datasets.}
We evaluate the proposed ArcAD on four standard industrial datasets. \emph{MVTec-AD}~\cite{bergmann2019mvtec} features 15 categories, comprising 3,629 normal training images and a test set of 1,725 images. \emph{VisA}~\cite{zou2022spot} consists of 12 distinct categories. The official data split provides 8,659 normal samples for training and 2,162 for testing. Furthermore, we utilize the large-scale \emph{Real-IAD}~\cite{wang2024real} dataset, which covers 30 industrial categories. There are 36,465 normal images for training and a test set of 114,585 images. \emph{MANTA}~\cite{fan2025manta} contains 38 object categories. We utilize the tiny version of this dataset, which comprises 107,415 images in the training set and 44,585 images in the testing set. 

\begin{figure*}[t]
    \centering
\includegraphics[width=1.0\linewidth]{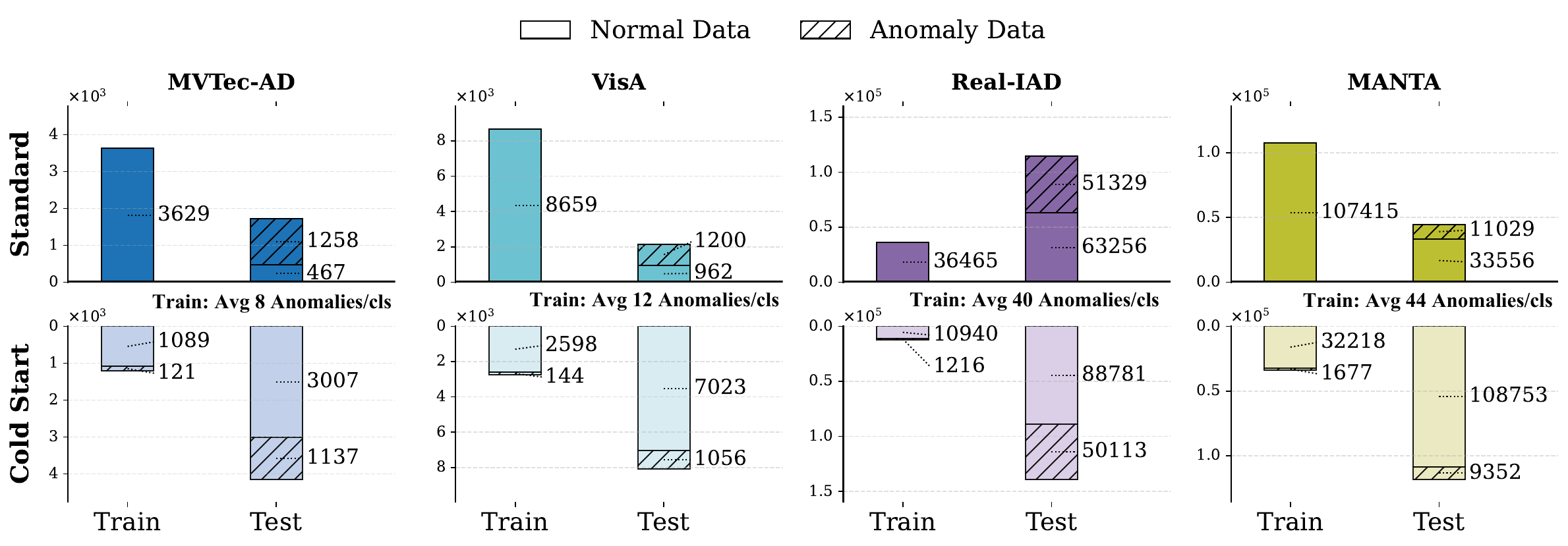} 
    \caption{Statistics of standard and cold-start setting datasets (see details in \textit{Supp}).}
    \label{fig:dataset}

\end{figure*}

In our cold-start setting, 30\% of the standard training set is utilized as normal training data, which is then combined with a subset of anomalies sampled from the original test set. Typically, anomalies account for 10\% of this newly constructed training set. Due to the severe scarcity of anomalous samples, we set this ratio to 5\% for VisA and MANTA. Unlike standard unsupervised datasets, the cold-start training set contains fewer normal samples and introduces a few anomalies. The test set incorporates the normal samples removed from the training set and contains reduced anomalies. Unsupervised methods are trained solely on the normal subset of the training data, whereas supervised methods and our approach utilize the full cold-start training set. All baselines are evaluated on this identical cold-start test set. Detailed statistics are presented in Fig.~\ref{fig:dataset}.

\paragraph{Evaluation Metrics.}
Consistent with established protocols~\cite{guo2025dinomaly,zhang2023exploring}, we evaluate performance using three primary metrics. For image-level anomaly detection, we report the Area Under the Receiver Operating Characteristic curve (I-AUROC). For pixel-level localization, we assess performance via P-AUROC and P-F$_1$-max. Additional metrics are provided in the \textit{Supp.} All reported scores are averaged across all classes.

\paragraph{Implementation Details.}
We integrate our proposed framework into three reconstruction based methods: Dinomaly~\cite{guo2025dinomaly}, RD4AD~\cite{deng2022anomaly}, and ReContrast~\cite{guo2023recontrast}. Data preprocessing follows the standard input and output configurations of each respective framework. All models are trained for $10,000$ iterations across all evaluated datasets. We optimize the networks using the AdamW optimizer. The hyperparameters are configured with a learning rate of $2 \times 10^{-3}$, momentum parameters $\beta = (0.9, 0.999)$, and weight decay of $1 \times 10^{-4}$. The balancing hyperparameters $\lambda_1, \lambda_2$, and $\lambda_3$ in our total loss function are set to 0.1. All baseline methods are reimplemented under the same cold-start setting. 


\subsection{Main Results}

\begin{table*}[t]
\centering
\caption{Performance under multi-class setting (\%). Each cell reports
I-AUROC / P-AUROC / P-F$_1$-max. Detailed per-class performances are presented in the \textit{Supp}.}
\label{tab:multiclass_performance}
\resizebox{\textwidth}{!}{
\begin{tabular}{llcccc}
\toprule
\textbf{} & \textbf{Method}
& \textbf{MVTec-AD}~\cite{bergmann2019mvtec}
& \textbf{VisA}~\cite{zou2022spot}
& \textbf{Real-IAD}~\cite{wang2024real}
& \textbf{MANTA}~\cite{fan2025manta} \\ 
\midrule

\multirow{7}{0.6cm}{\centering\rotatebox{90}{\textbf{Unsup.}}}
& SimpleNet~\cite{liu2023simplenet}  & 79.4/84.5/30.7 & 85.1/92.8/37.6 & 63.6/66.3/2.5 & 60.3/72.2/7.3 \\
& DeSTSeg~\cite{zhang2023destseg}    & 94.5/97.3/60.9 & 86.6/96.0/39.5 & 83.5/92.7/39.8 & 75.5/85.0/20.0 \\
& MambaAD~\cite{he2024mambaad}    & 95.5/96.7/56.8 & 94.2/97.6/39.1 & 85.3/97.7/38.2 & 85.2/91.8/29.8 \\
& ReContrast~\cite{guo2023recontrast} & 93.2/97.6/57.8 & 93.2/98.0/46.0 & 85.9/97.8/40.8 & 86.4/93.4/30.5 \\
& RD4AD~\cite{deng2022anomaly}      & 91.4/96.2/44.1 & 90.3/97.5/42.2 & 77.8/96.9/29.7 & 76.8/91.8/23.5 \\
& UniNet~\cite{wei2025uninet}      & 92.7/96.4/49.7 & 90.9/98.0/44.4 & 76.5/96.8/32.9 & 68.4/89.9/15.4 \\
& Dinomaly~\cite{guo2025dinomaly}   & 99.6/98.3/68.4 & 98.4/98.9/53.9 & 88.8/98.8/46.7 & 90.3/93.8/39.8 \\

& \cellcolor{gray!15}Dinomaly*~\cite{guo2025dinomaly} &\cellcolor{gray!15}99.6/98.4/69.2 & \cellcolor{gray!15}98.7/98.7/55.7 & \cellcolor{gray!15}89.3/98.8/47.1& \cellcolor{gray!15}91.1/95.1/53.1 \\
\midrule

\multirow{5}{0.6cm}{\centering\rotatebox{90}{\textbf{Sup.}}}
& SDNet~\cite{tabernik2020segmentation} & 70.1/49.4/6.9 &79.0/54.9/1.4& 74.7/63.3/0.9 &78.5/66.9/4.2  \\ 
& SDNet~\cite{tabernik2020segmentation} (Dinov2) & 59.7/60.2/3.9 &77.4/42.1/0.6&55.4/71.8/0.8 & 60.7/70.6/5.1\\ 
& DevNet~\cite{pang2021explainable}  & 57.7/61.5/3.7 & 50.5/51.6/3.9  & 50.0/59.2/7.1 & 51.3/49.7/6.1 \\
& DRA~\cite{ding2022catching}   & 85.2/63.9/13.1 & 79.3/49.9/1.7  & 76.0/60.3/4.9 & 77.4/34.8/1.9 \\
& \cellcolor{CornflowerBlue!10}ArcAD\textsuperscript{\textdagger}
& \cellcolor{CornflowerBlue!10}99.7/99.2/68.9
& \cellcolor{CornflowerBlue!10}98.9/99.0/54.9
& \cellcolor{CornflowerBlue!10}92.5/99.0/49.8
& \cellcolor{CornflowerBlue!10}93.3/95.5/48.5 \\
\bottomrule
\end{tabular}
}
\flushleft 
\footnotesize 
\textsuperscript{\textdagger} \textit{Baseline built upon Dinomaly.} \textsuperscript{*}\textit{Ref: performance under standard setting
from~\cite{guo2025one}.}
\end{table*}

\paragraph{Multi-class Performance.}
We conducted experiments under the widely applied multi-class setting, where all classes of samples are mixed for training. We compared the proposed ArcAD based on Dinomaly, with both unsupervised~\cite{liu2023simplenet,zhang2023destseg,he2024mambaad,guo2023recontrast,deng2022anomaly,wei2025uninet,guo2025dinomaly} and supervised methods~\cite{tabernik2020segmentation,pang2021explainable,ding2022catching}. Experimental results are presented in Table~\ref{tab:multiclass_performance}. Overall, the proposed ArcAD consistently outperforms other state-of-the-art (SOTA) methods across all datasets and metrics. 

Compared to unsupervised learning baselines, our approach yields significant improvements across the board. On both the MVTec-AD and VisA, ArcAD achieves the best performance with 99.7\%/99.2\%/68.9\% and 98.9\%/99.0\%/54.9\% (I-AUROC/P-AUROC/P-F$_1$-max), surpassing all SOTA methods. This performance gap widens significantly on more complex, large-scale datasets with over 30 categories. Specifically, on the Real-IAD dataset, ArcAD achieves an I-AUROC of 92.5\% and a P-F$_1$-max of 49.8\%, exceeding Dinomaly by notable margins of 3.7\% and 3.1\%. Similarly, on the MANTA dataset, ArcAD obtains an I-AUROC of 93.3\% and a P-F$_1$-max of 48.5\%, outperforming the best Dinomaly by 3.0\% and 8.7\%. Compared to unsupervised methods, our approach leverages a few anomaly samples as guidance to explicitly constrain and calibrate the normal manifold by SPM and DGC, thereby yielding a highly discriminative boundary. 

Furthermore, we compared ArcAD against previous supervised learning approaches. Additionally, we replaced the backbone of SDNet with the identical Dinov2~\cite{oquab2024dinov2} model used in our baseline. Specifically, we achieve an I-AUROC of 99.7\% on MVTec-AD (+14.5\% over DRA), 98.9\% on VisA (+19.6\% over DRA), 92.5\% on Real-IAD (+16.5\% over DRA) and 93.3\% on MANTA (+14.8\% over SDNet). This is primarily because conventional supervised models are highly prone to overfitting the scarce anomalous samples available in cold start scenarios. Compared to the other supervised methods, ArcAD leverages the limited anomaly data to rectify the normal distribution, achieving significantly more accurate anomaly detection and superior generalization capabilities.

\begin{table*}[t]
\centering
\caption{Performance under single-class setting (\%). Each cell reports
I-AUROC / P-AUROC / P-F$_1$-max.}
\label{tab:singleclass_performance}

\resizebox{\textwidth}{!}{
\begin{tabular}{llcccc}
\toprule
\textbf{} & \textbf{Method}
& \textbf{MVTec-AD}~\cite{bergmann2019mvtec}
& \textbf{VisA}~\cite{zou2022spot}
& \textbf{Real-IAD}~\cite{wang2024real}
& \textbf{MANTA}~\cite{fan2025manta} \\ 
\midrule

\multirow{4}{0.6cm}{\centering\rotatebox{90}{\textbf{Unsup.}}}
& SimpleNet~\cite{liu2023simplenet}  & 99.1/98.9/58.8 & 95.6/98.7/46.4 & 88.3/96.3/30.8 & 87.5/95.6/43.4 \\
& ReContrast~\cite{guo2023recontrast} & 98.4/98.3/62.8 & 93.2/96.9/41.7 & 90.7/98.9/44.0 & 92.8/96.2/35.0 \\
& RD4AD~\cite{deng2022anomaly}      & 96.6/98.1/60.0 & 94.5/98.3/45.5 & 90.0/98.8/45.8 & 89.7/94.7/32.8 \\
& Dinomaly~\cite{guo2025dinomaly}   & 99.7/99.0/66.8 & 98.6/98.9/52.6 & 92.0/98.9/47.1 & 94.3/96.2/43.7 \\
\midrule

\multirow{2}{0.6cm}{\centering\rotatebox{90}{\textbf{Sup.}}}
& DRA~\cite{ding2022catching}   & 76.3/67.8/9.8 & 89.1/57.7/8.1 & 81.1/82.4/15.1 & 90.6/48.6/4.3 \\
& \cellcolor{CornflowerBlue!10}ArcAD\textsuperscript{\textdagger}
& \cellcolor{CornflowerBlue!10}100.0/99.3/67.5
& \cellcolor{CornflowerBlue!10}99.1/99.1/52.7
& \cellcolor{CornflowerBlue!10}95.4/99.4/48.5
& \cellcolor{CornflowerBlue!10}96.0/97.2/48.0 \\
\bottomrule
\end{tabular}
}

\flushleft 
\footnotesize 
\textsuperscript{\textdagger} \textit{Baseline built upon Dinomaly.}
\end{table*}

\paragraph{Single-class Performance.} We evaluated ArcAD under the single-class setting across four datasets, as summarized in Table~\ref{tab:singleclass_performance}. Overall, ArcAD achieves the best overall anomaly detection performance. We achieve 100.0\%/99.3\%/67.5\% (I-AUROC/P-AUROC/P-F$_1$-max) on MVTec-AD, outperforming the best competing method by (+0.3\%/+0.3\%/+0.7\%). On VisA dataset, our approach obtains 99.1\%/99.1\%/52.7\%, exceeding Dinomaly by (+0.5\%/+0.2\%/+0.1\%). The advantages are even more pronounced on challenging large-scale datasets. On Real-IAD, ArcAD reaches 95.4\%/99.4\%/48.5\%, surpassing the top competitor by notable margins of (+3.4\%/+0.5\%/+1.4\%). Similarly, on the MANTA dataset, ArcAD obtains 96.0\%/97.2\%/48.0\%, yielding significant improvements of (+1.7\%/+1.0\%/+4.3\%) over the best Dinomaly. These results demonstrate that ArcAD remains highly effective under the single-class setting.

\begin{figure*}[t] 
    \centering
    \includegraphics[width=1.0\textwidth]{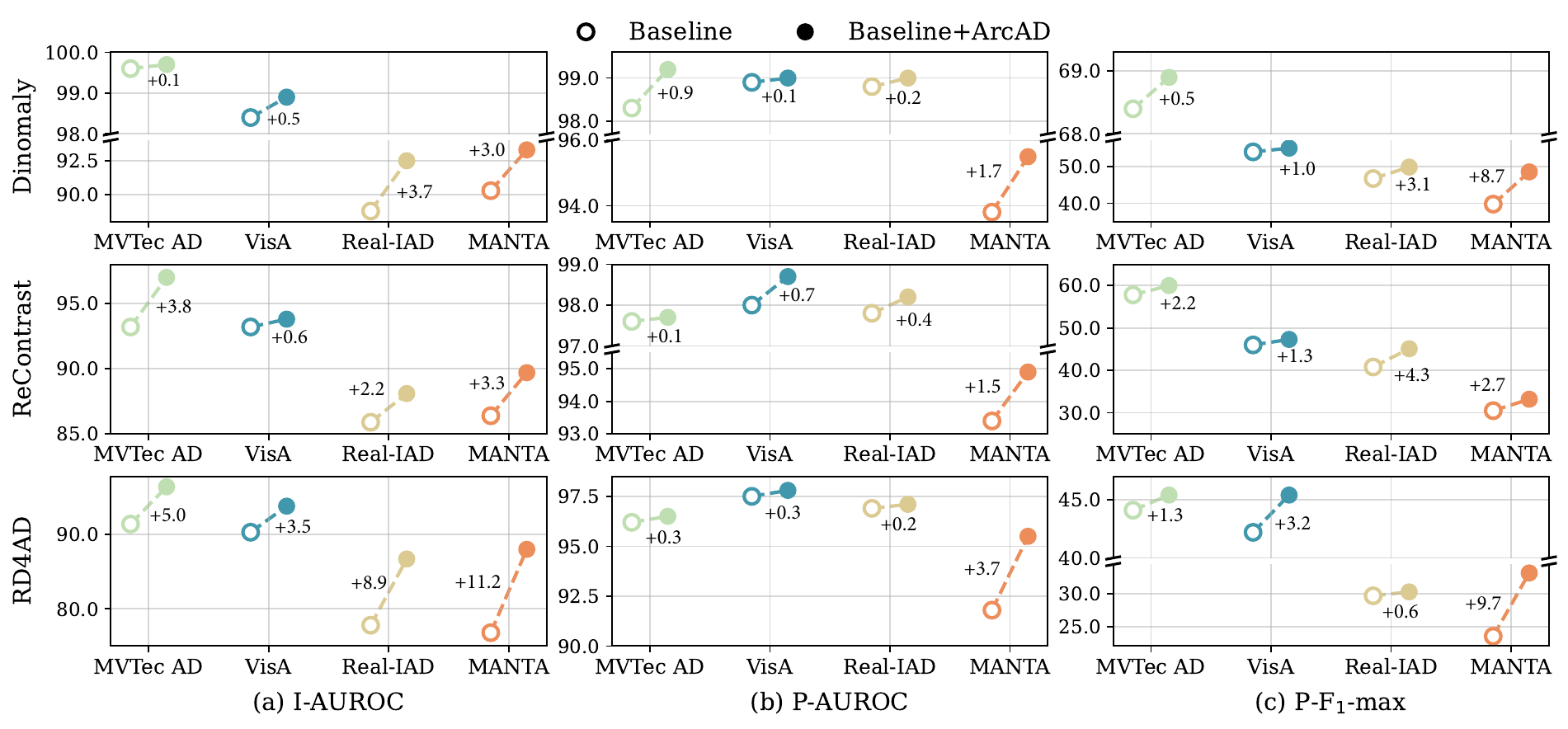}
    \caption{
        Performance of baseline methods (Dinomaly, ReContrast, and RD4AD) equipped with the ArcAD across four datasets under the multi-class setting.
    }
    \label{fig:baseline}

\end{figure*}

\paragraph{Generality of ArcAD.}
To validate the generality of our proposed ArcAD, we integrated it into three mainstream reconstruction-based anomaly detection baselines (Dinomaly, ReContrast, and RD4AD) and evaluated them across four datasets under the multi-class setting. As illustrated in Fig.~\ref{fig:baseline}, after incorporating the ArcAD framework, all baseline models exhibit improvements across all metrics on the four datasets. Specifically, integrating ArcAD into the RD4AD framework yields I-AUROC scores of 96.4\% (+5.0\%), 93.8\% (+3.5\%), 86.7\% (+8.9\%), and 88.0\% (+11.2\%) on MVTec-AD, VisA, Real-IAD, and MANTA, respectively. This demonstrates that ArcAD does not overfit to specific methods but rather fundamentally enhances the discriminative capability of the reconstruction-based models in cold-start scenarios.

\paragraph{Effect of Anomaly Ratio.} To evaluate the sensitivity of ArcAD to varying anomaly ratios, we conducted experiments with different anomaly ratios (3\%, 5\%, and 10\%) on the Real-IAD dataset under the multi-class setting. As reported in Table~\ref{tab:anomaly_ratio}, the integration of ArcAD consistently yields performance improvements across all three baseline models. Specifically, at a 10\% anomaly ratio, ArcAD boosts the cold-start I-AUROC of Dinomaly, ReContrast, and RD4AD to 92.5\% (+3.7\%), 88.1\% (+2.2\%), and 86.7\% (+8.9\%), respectively. Remarkably, even at a mere 3\% anomaly ratio, our method still improves the I-AUROC of Dinomaly, ReContrast, and RD4AD by 0.9\%, 2.2\%, and 5.2\%, respectively. The ArcAD serves as a highly generalizable enhancement, consistently elevating the performance regardless of the anomaly ratio.
\definecolor{myblue}{HTML}{005DB3}
\definecolor{bestcolor}{HTML}{51B5A1}   
\definecolor{secondcolor}{HTML}{89C3B9} 
\definecolor{thirdcolor}{HTML}{CDE9E4}  

\begin{table*}[t]
\centering
\renewcommand{\arraystretch}{1.5}
\caption{Performance comparison (I-AUROC / P-AUROC / P-F$_1$-max) under different anomaly ratios on Real-IAD dataset. Subscripts denote the performance improvement compared to the cold-start performance.}
\label{tab:anomaly_ratio}
\resizebox{\textwidth}{!}{
\begin{tabular}{l|c|c >{\columncolor{thirdcolor!50}}c >{\columncolor{secondcolor!50}}c >{\columncolor{bestcolor!50}}c}
\toprule
\textbf{Method} & \textbf{Official} & \textbf{Cold Start} & 
\multicolumn{1}{c}{\textbf{+ArcAD 3\%}} & 
\multicolumn{1}{c}{\textbf{+ArcAD 5\%}} & 
\multicolumn{1}{c}{\textbf{+ArcAD 10\%}} \\
\midrule
ReContrast & 86.4/97.8/38.2$^*$ & 85.9/97.8/40.8
& $88.1_{\mathbf{+2.2}}$/$98.0_{\mathbf{+0.2}}$/$44.9_{\mathbf{+4.1}}$ 
& $88.2_{\mathbf{+2.3}}$/$98.2_{\mathbf{+0.4}}$/$45.2_{\mathbf{+4.4}}$ 
& $88.1_{\mathbf{+2.2}}$/$98.2_{\mathbf{+0.4}}$/$45.1_{\mathbf{+4.3}}$ \\

RD4AD & 82.4/97.3/32.7$^*$ & 77.8/96.9/29.7
& $83.0_{\mathbf{+5.2}}$/$97.1_{\mathbf{+0.2}}$/$33.0_{\mathbf{+3.3}}$ 
& $84.4_{\mathbf{+6.6}}$/$97.0_{\mathbf{+0.1}}$/$33.1_{\mathbf{+3.4}}$ 
& $86.7_{\mathbf{+8.9}}$/$97.1_{\mathbf{+0.2}}$/$30.3_{\mathbf{+0.6}}$ \\

Dinomaly & 89.3/98.8/47.7$^*$ & 88.8/98.8/46.7
& $89.7_{\mathbf{+0.9}}$/$98.8_{\mathbf{+0.0}}$/$48.6_{\mathbf{+1.9}}$ 
& $90.8_{\mathbf{+2.0}}$/$98.8_{\mathbf{+0.0}}$/$49.4_{\mathbf{+2.7}}$ 
& $92.5_{\mathbf{+3.7}}$/$99.0_{\mathbf{+0.2}}$/$49.8_{\mathbf{+3.1}}$ \\
\bottomrule
\end{tabular}
}
\flushleft
\footnotesize
 $^*$ \textit{Ref: performance under standard setting
from~\cite{guo2025dinomaly}.}
\end{table*}

\subsection{Ablation Study}
\paragraph{Overall Ablation.} To investigate the individual contribution of each module, we conducted an incremental ablation study on the Real-IAD dataset under the multi-class setting. The results are shown in Table~\ref{tab:ablation_components}. First, removing the SPM module ($\mathcal{L}_{spm}$) decreases I-AUROC, P-AUROC, and P-F$_1$-max by 0.8\%, 0.1\%, and 0.5\%, respectively, verifying the necessity of constraining feature uniformity and compactness. Second, removing the DGC module ($\mathcal{L}_{dgc}$) drops the metrics by 2.3\%, 0.2\%, and 1.7\%, demonstrating that anomaly-guided distribution calibration is crucial for rectifying the normal boundary.

\begin{wraptable}[11]{r}{0.54\textwidth} 
    \centering
    \caption{Ablation of individual components.}
    \label{tab:ablation_components}
    
    \resizebox{\linewidth}{!}{
        \begin{tabular}{c c c c | c | c c}
            \toprule
            $\mathcal{L}_{recon}$ & $\mathcal{L}_{spm}$ & $\mathcal{L}_{cls}$ & $\mathcal{L}_{dgc}$ & I-AUROC & P-AUROC & P-F$_1$-max \\
            \midrule
            \checkmark & & & & 88.8&98.8&46.7  \\
            \checkmark & \checkmark & & \checkmark & 90.2 & 98.9 & 49.1 \\
            \checkmark &  & \checkmark & \checkmark &  91.7 & 98.9 & 49.3 \\
            \checkmark & \checkmark & \checkmark & & 90.2 & 98.8 & 48.1 \\
            \checkmark & & \checkmark & & 87.8 & 98.5 & 45.5 \\
            \checkmark & \checkmark & \checkmark & \checkmark & \textbf{92.5} & \textbf{99.0} & \textbf{49.8} \\
            \bottomrule
        \end{tabular}
    }
\end{wraptable}
Furthermore, eliminating the discriminator loss ($\mathcal{L}_{cls}$) drops performance by 2.3\%, 0.1\%, and 0.7\%, confirming its effectiveness in boundary refinement. Notably, adding only the discriminator to the baseline actually degrades performance (\textit{e.g.,} I-AUROC drops from 88.8\% to 87.8\%), proving its effectiveness intrinsically relies on synergy with the SPM and DGC.

\paragraph{Clustering Strategy.}
To investigate the effectiveness of the SPM, we replace the proposed clustering strategy with Euclidean distance-based $K$-means and cosine distance-based $K$-means, respectively. As shown in Table~\ref{tab:ablation_clustering}, the Sinkhorn-based clustering achieves the highest performance of 92.5\%/99.0\%/49.8\%, outperforming $K$-means (Euclidean) by margins of 20.1\%/6.2\%/24.0\% and surpassing $K$-means (Spherical) by 1.6\%/0.1\%/3.3\%.

\begin{wraptable}{r}{0.51\textwidth}
    \centering
    
    \caption{Ablation of prototyping strategies.}
    \label{tab:ablation_clustering}

    \resizebox{0.5\textwidth}{!}{
    \begin{tabular}{l | c | c c}
        \toprule
         Prototyping Method & I-AUROC & P-AUROC & P-F$_1$-max \\
        \midrule
        $K$-means (Euclidean) & 72.4 & 92.8 & 25.8 \\
        $K$-means (Spherical) & 90.9 & 98.9 & 46.5 \\
        \textbf{Sinkhorn (Ours)} & \textbf{92.5} & \textbf{99.0} & \textbf{49.8} \\
        \bottomrule
    \end{tabular}}
    
\end{wraptable}
In particular, the substantial improvement over $K$-means (Euclidean) demonstrates the necessity of optimizing on the hypersphere to prevent mode collapse. Compared to the biased cluster assignments often produced by spherical $K$-means, the proposed SPM enforces an equipartition constraint, thereby achieving a more uniform and compact manifold.

\paragraph{Data Augmentation.} To evaluate the effectiveness of our proposed anomaly synthesis strategy, we compared it against variants. As reported in Table~\ref{tab:ablation_aug}, our prototype-restricted anomaly synthesis strategy achieves the best performance of 92.5\%/99.0\%/49.8\%. It improves upon the baseline without any anomaly synthesis by 0.9\%/0.2\%/0.7\%, and surpasses the augmentation strategies applied in SimpleNet and DRAEM by 3.6\%/0.4\%/2.0\% and 4.1\%/0.3\%/2.8\%, respectively.
\begin{wraptable}{rt}{0.6\textwidth}
    \centering
    \caption{Ablation of anomaly synthesis strategies.}
    \label{tab:ablation_aug}
    \resizebox{0.6\textwidth}{!}{
        \begin{tabular}{l | c | c c}
            \toprule
            \textbf{Synthesis Strategy}& I-AUROC & P-AUROC & P-F$_1$-max  \\
            \midrule
            w/o Anomaly synthesis & 91.6 & 98.8 & 49.1 \\
            w/o Filter & 90.2 & 98.5 & 43.4 \\    SimpleNet~\cite{liu2023simplenet} & 88.9 & 98.6 & 47.8 \\          DRAEM~\cite{zavrtanik2021draem} & 88.4 & 98.7 & 47.0 \\
            \textbf{Ours} & \textbf{92.5} & \textbf{99.0} & \textbf{49.8} \\
            \bottomrule
        \end{tabular}
    } 

\end{wraptable}
Notably, compared to the strategy without filter (w/o Filter), which applies to all candidate anomalies, ArcAD yields a significant gain of 2.3\%/0.5\%/6.4\%. This improvement demonstrates that utilizes all candidate anomalies is sub-optimal. By selecting pseudo-anomalies far from normal prototypes, our strategy yields hard anomalies, enabling accurate boundary delineation.

\begin{wrapfigure}{r}{0.48\textwidth} %
    \centering
    
\includegraphics[width=1.0\linewidth]{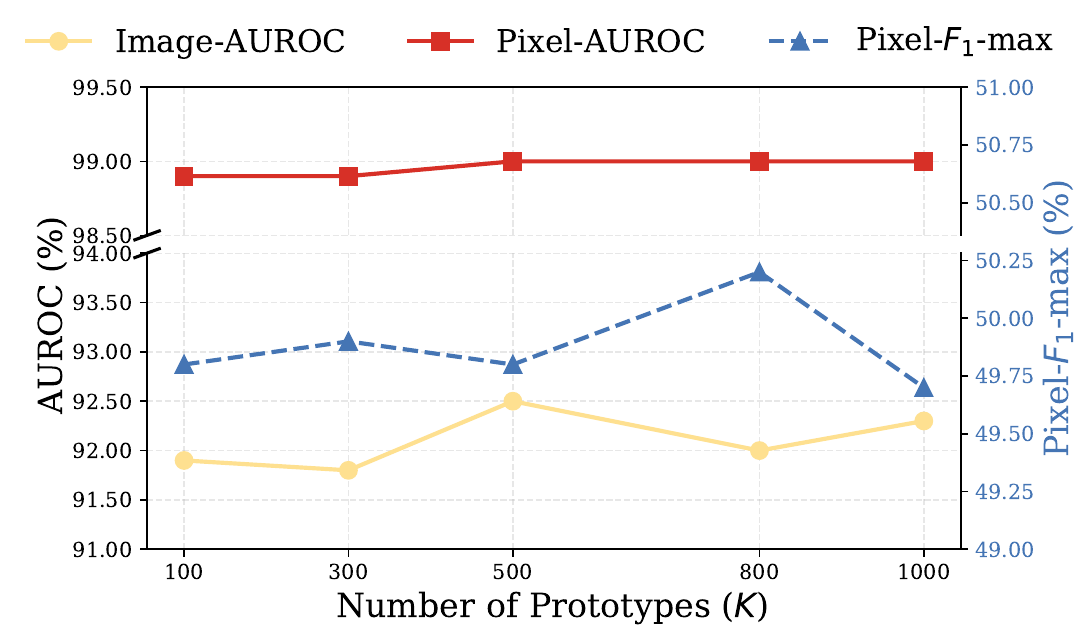} 
    \caption{Ablation of the number of prototypes $K$ in the SPM.}
    \label{fig:protos}
\end{wrapfigure}

\paragraph{Number of Prototypes $K$.} To evaluate the impact of the prototype quantity in the SPM module, we conducted an ablation study by varying the number of prototypes $K$ across $\{100, 300, 500, 800, 1000\}$. The model reaches its highest I-AUROC (92.5\%) at $K=500$, with P-AUROC staying relatively stable above 99.0\%. Although the highest P-F$_1$-max (50.2\%) occurs at $K=800$, we ultimately select $K=500$ as it provides the best balance between image-level and pixel-level detection performance.

\subsection{Visualization}
         




\begin{figure*}[tb]
  \centering
  \begin{minipage}[t]{0.48\linewidth}
    \centering
    \includegraphics[width=\linewidth]{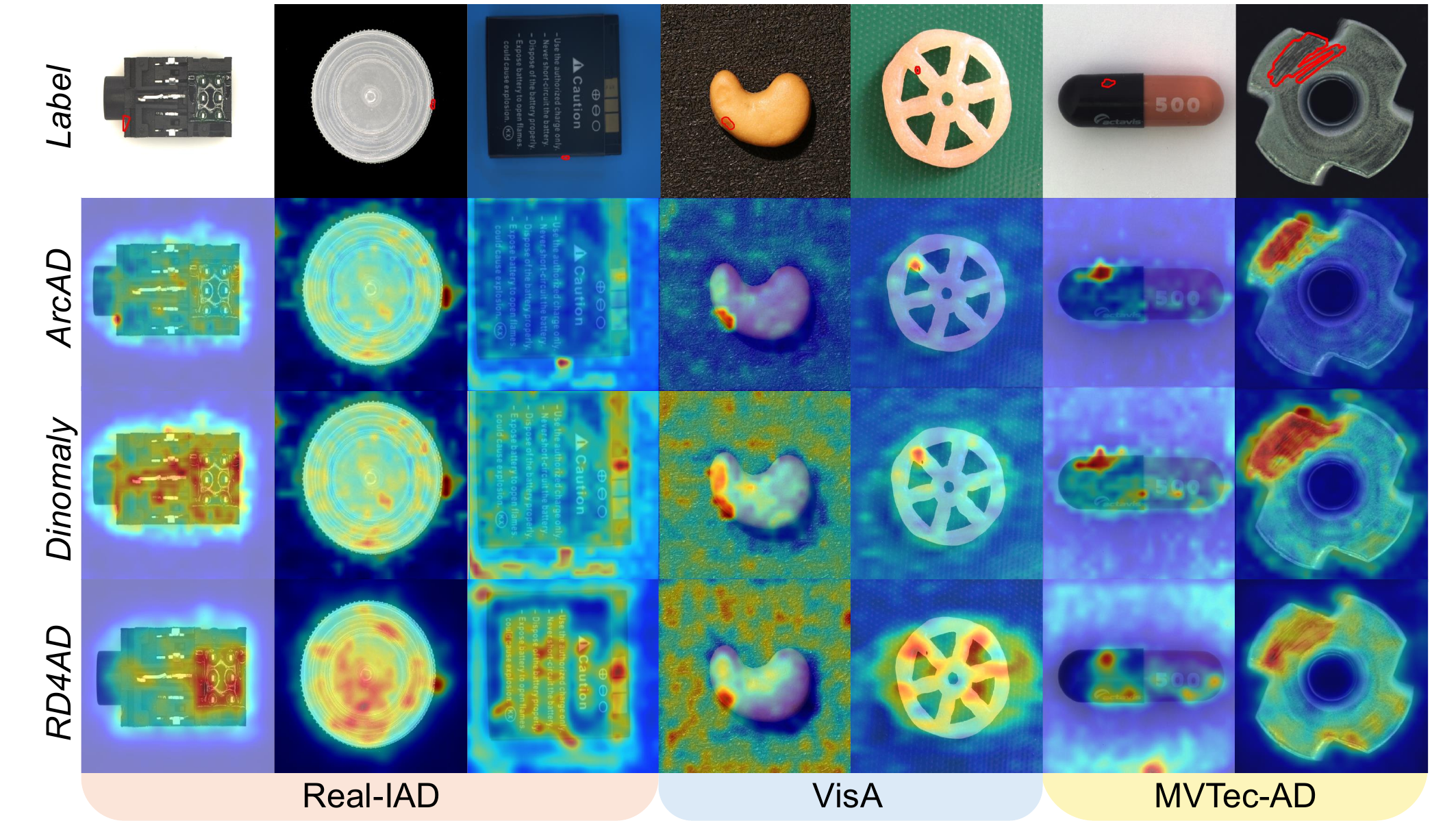}
  \end{minipage}\hfill
  \begin{minipage}[t]{0.48\linewidth}
    \centering
    \includegraphics[width=\linewidth]{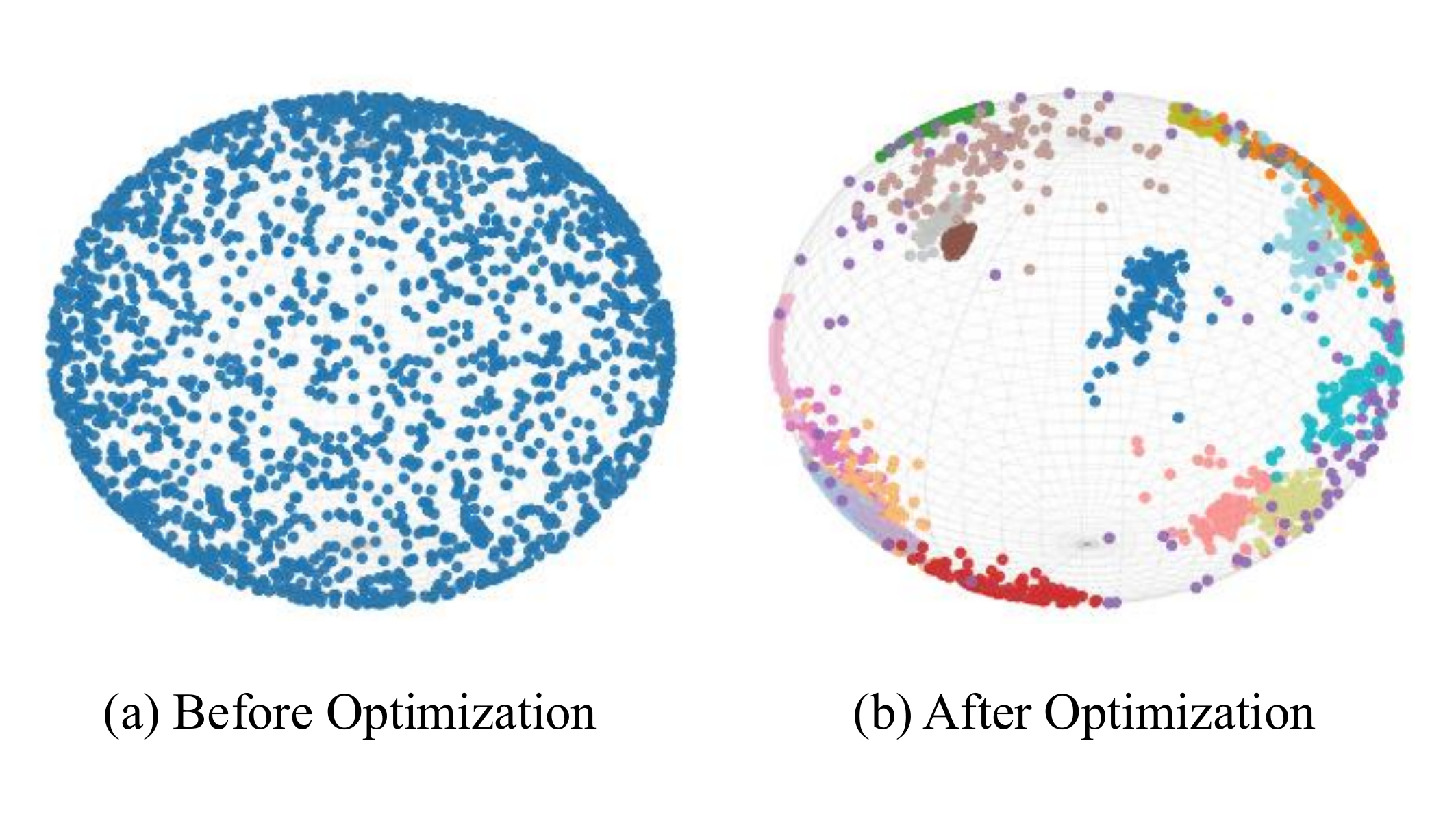}
  \end{minipage}
  
  \caption{\textbf{Left:} Qualitative results of anomaly localization for multi-class anomaly detection. Test images with anomalous regions are highlighted in red contours. \textbf{Right:} Feature visualization of ArcAD before and after optimization on the MANTA dataset.}
  \label{fig:visualization}
\end{figure*}

\paragraph{Pixel-level Localization.} We qualitatively visualize the localization results of our proposed ArcAD (based on Dinomaly), as illustrated in Fig.~\ref{fig:visualization} (left). Compared to the original Dinomaly and RD4AD baselines, incorporating ArcAD significantly reduces both false positives and false negatives, thereby yielding superior segmentation performance, especially on challenging small-scale anomalies. By enforcing a compact and uniform normal feature distribution while calibrated by anomalies, the model establishes a highly discriminative decision boundary.




\paragraph{Feature Visualization.} We visualize the learned representations on a 3D spherical manifold using t-SNE with cosine similarity applied to the $\ell_2$-normalized normal patch features extracted from the bottleneck. Specifically, we randomly select 20 clusters from the MANTA dataset under the multi-class setting to compare the feature distributions before and after optimization. As illustrated in Fig.~\ref{fig:visualization} (right), before optimization, the patch features are randomly and sparsely scattered across the hypersphere. In contrast, after applying our proposed optimization, the representations are effectively constrained, forming highly uniform and compact clusters.

\section{Conclusion}
We propose ArcAD, a novel plug-and-play calibration framework that effectively overcomes the cold-start bottleneck in IAD. By integrating Sinkhorn-based prototype modeling with defect-guided calibration, ArcAD explicitly rectifies the normal manifold. Extensive experiments across four datasets demonstrate that ArcAD achieves SOTA performance in both single-class and multi-class settings.

\paragraph{Limitations.} Although our approach introduces additional computational cost during the training phase, it does not compromise inference efficiency due to its direct reliance on standard reconstruction discrepancy. Future work will explore more lightweight prototype optimization and calibration strategies to further accelerate the training process.

\section*{Acknowledgements}
This work was supported by the National Natural Science Foundation of China (Grant No. 62277011), CAAI-CANN Open Fund, developed on OpenI Community, and Project of Chongqing MEITC (Grant No. YJX-2025001001009).

%
%
\bibliographystyle{splncs04}
\bibliography{main}
\end{document}